\documentclass[runningheads]{llncs}

\usepackage[T1]{fontenc}
\usepackage{graphicx}
\usepackage{multirow}
\usepackage{array}
\usepackage{booktabs} 
\usepackage{amsmath}
\usepackage{amssymb, bm}
\usepackage{makecell}
\usepackage[flushleft]{threeparttable}
\usepackage[title]{appendix}
\usepackage{tikz}
\usetikzlibrary{matrix}
\usepackage{subcaption}
\usepackage{nicefrac, xfrac}

\usepackage{hyperref}
\usepackage{color}

\usepackage{xspace}
\newcommand{\longname}{Abductive Rule Learner with Context-awareness\xspace}
\newcommand{\name}{ARLC\xspace}

\begin{document}

\title{Towards Learning Abductive Reasoning \\ using VSA Distributed Representations}
\titlerunning{Towards Learning Abductive Reasoning using VSA Dist. Representations}
\author{
    Giacomo Camposampiero\inst{1,2} \and
    Michael Hersche\inst{1}\and
    Aleksandar Terzić\inst{1,2} \and \\
    Roger Wattenhofer\inst{2} \and
    Abu Sebastian\inst{1} \and
    Abbas Rahimi\inst{1}
}
\authorrunning{Camposampiero et al.}

\institute{
    IBM Research -- Zurich \and
    ETH Zürich \\ 
    \email{giacomo.camposampiero1@ibm.com}
}

\maketitle

\newcolumntype{P}[1]{>{\centering\arraybackslash}p{#1}}
\renewcommand{\arraystretch}{1.4}

\begin{abstract}
We introduce the \longname (\name), a model that solves abstract reasoning tasks based on Learn-VRF.
\name features a novel and more broadly applicable training objective for abductive reasoning, resulting in better interpretability and higher accuracy when solving Raven's progressive matrices (RPM).
\name allows both programming domain knowledge and learning the rules underlying a data distribution.
We evaluate \name on the I-RAVEN dataset, showcasing state-of-the-art accuracy across both in-distribution and out-of-distribution (unseen attribute-rule pairs) tests.
\name surpasses neuro-symbolic and connectionist baselines, including large language models, despite having orders of magnitude fewer parameters.
We show \name's robustness to post-programming training by incrementally learning from examples on top of programmed knowledge, which only improves its performance and does not result in catastrophic forgetting of the programmed solution.
We validate \name's seamless transfer learning from a 2x2 RPM constellation to unseen constellations.
Our code is available at \href{https://github.com/IBM/abductive-rule-learner-with-context-awareness}{https://github.com/IBM/abductive-rule-learner-with-context-awareness}.
\end{abstract}

\section{Introduction}
\let\thefootnote\relax\footnotetext{\footnotesize\noindent 18th Conference on Neural-Symbolic Learning and Reasoning (NeSy 2024).}
Abstract reasoning can be defined as the ability to induce rules or patterns from a limited source of experience and generalize their application to similar but unseen situations.
It is widely acknowledged as a hallmark of human intelligence, and great efforts have been poured into the challenge of endowing artificial intelligence (AI) models with such capability.

As a result, a wide range of benchmarks to assess human-like fluid intelligence and abstract reasoning in AI models has been proposed in the past decade~\cite{RavenTest2012,cherian2023smart,chollet2019measure,niedermayr2023rlp}.
In this work, we focus on Raven’s progressive matrices (RPM) test~\cite{RavenTest2012,Carpenter1990,Raven1938}.
RPM is a visual task that involves perceiving pattern continuation and elemental abstraction as well as deducing relations based on a restricted set of underlying rules, in a process that mirrors the attributes of advanced human intelligence~\cite{snow1984topography,snow1984toward}.
Recently, RPM has become a widely used benchmark for effectively testing AI capabilities in abstract reasoning, making analogies, and dealing with out-of-distribution (OOD) data~\cite{MRNet_CVPR2021,I-Raven,RPM_Survey2022,Mitchel_Survey_2021,Raven_19}.

With the advent of large language models, it was suggested that the attainment of abstract reasoning abilities required to solve this kind of task may hinge upon the scale of the model.
To support this claim, it was shown that adequately large pre-trained language models can exhibit emergent abilities for logical~\cite{wei2022emergent} and analogical~\cite{hu-etal-2023-context,webb2023emergent} reasoning.
Nevertheless, the internal mechanisms underlying the emergence of these abilities are still not well understood. 
In addition, recent works provided evidence on the acute brittleness of these abilities~\cite{gendron2024large,wu2024reasoning}, while others showed that language models fail to attain levels of general abstract reasoning comparable to humans~\cite{10208934,lewis2024using,odouard2022evaluating,thomm2024limits}.

An alternative and promising direction is neuro-symbolic AI.
Neuro-symbolic approaches combine sub-symbolic perception with various forms of symbolic reasoning, resulting in cutting-edge performance across a spectrum of domains, including visual~\cite{NS_MetaConcept_NIPS19,NS_ConceptLearner_ICLR19,Falcon_ICLR2022,NS-VQA_NIPS18}, natural language~\cite{Learn2reason_TPR}, causal~\cite{CLEVRER_ICLR2020}, mathematical~\cite{TP-transformer_2019}, and analogical~\cite{yang2022conceptual,nesy2022_knowledge,zhao2023interpretable,PrAE_CVPR21,hersche2023neuro} reasoning tasks.
In the context of RPM, recent neuro-symbolic architectures focused on abductive reasoning~\cite{PrAE_CVPR21,hersche2023neuro}. 
Abductive reasoning allows to selectively infer propositions based on prior knowledge represented in a symbolic form to explain the perceptual observations in the best possible way~\cite{AbductiveCognition}.
The appeal of the abductive approach lies in its accommodation of perceptual uncertainties within symbolic reasoning. 
Abductive reasoning can be implemented in systems that leverage distributed vector-symbolic architectures (VSAs)~\cite{VSA_03,Kanerva2009,PlateHolographic1995} representations and operators, such as the Neuro-Vector Symbolic Architecture (NVSA)~\cite{hersche2023neuro}.
However, these neuro-symbolic architectures~\cite{PrAE_CVPR21,hersche2023neuro} necessitate complete knowledge of the application domain (which might not be available) to program the right inductive bias into the model.

Learn-VRF~\cite{hersche2023probabilistic} overcomes this limitation by introducing a probabilistic abduction reasoning approach that learns a subset of the rules underlying RPM from data.
Learn-VRF transparently operates in the rule space, learning them through a soft assignment of VSA attribute representations to a fixed rule template.
During inference, it generates the answer panel by executing all the learned rules and applying a soft-selection mechanism to their outputs.
Nevertheless, Learn-VRF comes with several limitations, including a sub-optimal selection mechanism, poor performance on the RPM constellations involving multiple objects, and a constraint on the expressiveness of the RPM rules it can learn.

\begin{figure}[h!]
    \centering
    \includegraphics[width=\linewidth]{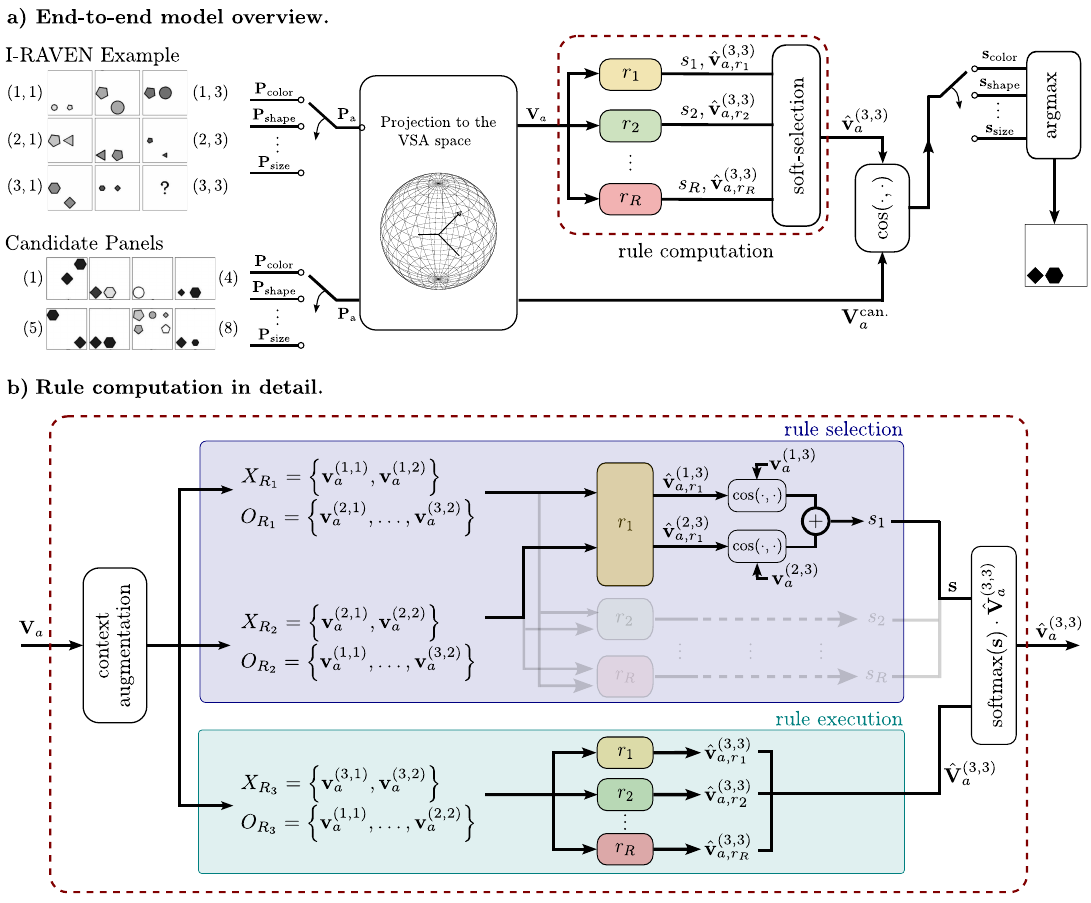}
    \caption{Proposed \name architecture. \textbf{a)} Overview of the end-to-end inference pipeline. \textbf{b)} Detailed rule computation block, exploded from (a), which highlights the difference between the two steps of the rule computation: rule selection and rule execution. We also illustrate the proposed context-augmentation abstraction (that translates VSA vectors into different sets $X_{R_i}$ and $O_{R_i}$ depending on the row ${R_i}$), and the parameter sharing between rules ($r_1, \dots, r_R$) across the selection and execution steps (rule blocks with the same color share the same parameters $\mathbf{w}$, $\mathbf{u}$, $\mathbf{v}$).}
    \label{fig:carl}
\end{figure}

To make progress towards learning-to-reason, we propose the \longname (\name) to tackle the main limitations of Learn-VRF~\cite{hersche2023probabilistic}. We advance a novel context-augmented formulation of the optimization problem and a more expressive rule template, which allows sharing rules with the same parameters in both execution and selection steps and offers better interpretability.
An overview of \name is depicted in Figure \ref{fig:carl}.
\name features programmability and can further learn from data on top of programmed knowledge.  
We evaluate \name on in-distribution (ID) and out-of-distribution (OOD) tests of the I-RAVEN dataset and demonstrate that \name significantly outperforms neuro-symbolic and connectionist baselines, including large language models. 
Further, the number of trainable parameters is reduced by two orders of magnitude compared to Learn-VRF.
We experimentally validate the programmability of \name by encoding domain knowledge, and discover that post-programming training, contrary to other studies~\cite{wu2023numerosity}, does not compromise the validity of the solution, but rather improves it.
Finally, training the model on a single constellation and evaluating it on all the others, we show that, unlike previous baselines~\cite{hersche2023probabilistic,hersche2023neuro}, the learned rules can seamlessly be transferred across constellations of the I-RAVEN dataset.

\section{Background}

\subsection{Vector Symbolic Architectures}

Vector-symbolic architectures (VSAs)~\cite{VSA_03,Kanerva2009,PlateHolographic1995} are a family of computational models that rely on the mathematical properties of high-dimensional vector spaces.
VSAs make use of high-dimensional distributed representations for structured (symbolic) representation of data while maintaining the advantages of connectionist distributed vector representations (see~\cite{VSA_Survey_Part1} for a survey).
Here is a formal definition of VSAs:

\begin{definition}[VSA]
\label{def:vsa}
A vector-symbolic architecture (VSA) consists of a 4-tuple $\mathbb{V}=(\mathbb{C}, \oplus, \otimes,\odot)$, where $\mathbb{C}$ is a set of high-dimensional distributed vectors equipped with two main operations, $\oplus$ (bundling) and $\otimes$ (binding), and on which it is possible to define a similarity measure $\odot$.
\end{definition}
Bundling is a similarity-preserving operation that creates a superposition of the operands, that is, the resulting vector will have a high similarity with the two operands.
Binding, on the other hand, is an operation that allows to bind a vector (value) to another vector (key) and does not preserve similarities; it usually allows an inverse operation, called unbinding.
The specific realization of the bundling, binding, and vector space constitutes the main difference between members of the VSA family.

\subsection{Raven's Progressive Matrices}
\label{sec:raven}
In this work, we focus on the I-RAVEN dataset~\cite{I-Raven}, a benchmark that provides RPM tests sampled from unbiased candidate sets to avoid short-cut solutions that were possible in the original RAVEN dataset~\cite{Raven_19}. 
Each RPM test is an analogy problem presented as a $3\times 3$ pictorial matrix of context panels. 
Every panel in the matrix is filled with several geometric objects based on a certain rule, except the bottom-right panel, which is left blank.
Figure~\ref{fig:carl} includes an I-RAVEN example test. 
The task is to complete the missing panel by picking the correct answer from a set of (eight) candidate answer panels that matches the implicit generation rule on every attribute. 
The object's attributes (color, size, shape, number, position) are governed by individual underlying rules: 
\begin{itemize}
    \item \textit{constant}, the attribute value does not change per row;
    \item \textit{arithmetic}, the attribute value of the third panel corresponds to either the sum or the difference of the first two panels of the row;
    \item \textit{progression}, the attribute value monotonically increases or decreases in a row by 1 or 2;
    \item \textit{distribute three}, the set of the three different values remains constant across rows, but the individual attribute values get shifted to the left or to the right by one position at every row; it also holds column-wise.
\end{itemize}
Each panel contains a variable number of objects (minimum one, maximum nine) arranged according to one of seven different constellations (center, distribute-four, distribute-nine, 
left-right, up-down, in-out-center, and in-out-four).

\subsection{Learning to Reason with Distributed Representations}
In this section, we discuss how vector-symbolic architectures can be used to solve tasks that require analogical and relational reasoning such as RPM.
In particular, we focus on Learn-VRF~\cite{hersche2023probabilistic}, a simple yet powerful approach that enables solving RPM tests by learning the underlying relations between visual attributes in the VSA representational space.

The key observation behind this approach is that the formulation of every RPM rule in the VSA algebra is a particular instance of a general rule template, which is shared among all the rules and consists of a series of binding and unbinding operations between VSA vectors.
Hence, the problem of learning RPM rules can be framed as an assignment problem between vectors representing visual attributes and terms in this general rule template.
This alternative formulation allows to tackle one of the main limitations of neuro-symbolic approaches, differentiability, and therefore enables the use of data-driven learning algorithms based on gradient optimization.

Learn-VRF includes several sequential steps, ranging from the translation of visual attributes into the VSA high-dimensional space to the computation of final results, which are detailed in the following paragraphs.

\subsubsection{From Visual Attributes to VSA.}
Following the same procedure of previous works which assume a perfect perception~\cite{hu2023context,Webb2023}, the panel's attribute labels are provided directly by the I-RAVEN metadata.
For every attribute $a$, each panel's label is translated to a probability mass function (PMF) $\mathbf{p}_a^{(i,j)}$, where $i$ is the row index and $j$ is the column index of the panel.
The panel's PMF is then projected into the VSA space as
\begin{align*}
    \mathbf{v}_a^{(i,j)} = \sum_{k=1}^N \mathbf{p}_a^{(i,j)}\left[k\right] \cdot \mathbf{b}\left[k\right], 
\end{align*}
where $N$ is the number of possible values that the attribute $a$ can assume.

The VSA vectors are drawn from a dictionary of binary generalized sparse block codes (GSBCs)~\cite{hersche2023factorizers}  $\mathbb{C}=\{\mathbf{b}_i \}_{i=1}^{512}$.
In binary GSBCs, the $D$-dimensional vectors are divided into $B$ blocks of equal length, $L=D/B$, where only one (randomly selected) element per block is set to 1 ($D=1024$ and $B=4$).

The algebraic operations on binary GSBCs are defined in Table \ref{tab:vecop}.
Combining GSBCs with fractional power encoding (FPE)~\cite{PlateHolographic2003} allows the representation of continuous attributes (e.g., color or size) and simple algebraic operations, as addition and subtractions, in the corresponding vector space.
In other words, the FPE initialization allows to establish a semantic equivalence between high-dimensional vectors and real numbers.
This property is consistently exploited in the framework, as it allows to solve the analogies in the puzzles as simple algebraic operations in the domain of real numbers.
Finally, we observe that the binding operation for binary GSBCs has properties analogous to addition in the real number domain, including commutativity, associativity, and the existence of a neutral element ($\mathbf{e} \in \mathbb{C}$, s.t. $\mathbf{a}\circledast \mathbf{e} = \mathbf{a}$ $\forall \mathbf{a} \in \mathbb{C}$ ).

\begin{table}[h!]
\caption{Supported VSA operations and their equivalent in $\mathbb{R}$.} 
\label{tab:vecop}
\centering
\begin{tabular}{ P{3cm} P{5cm} P{3cm} } 
\toprule
Operation & Binary GSBCs & Equivalent in $\mathbb{R}$ \\
\cmidrule(r){1-1}\cmidrule(r){2-2}\cmidrule(r){3-3}
Binding ($\otimes$) & Block-wise circular convolution ($\circledast$) & Addition $+$\\
Unbinding ($\oslash$) & Block-wise circular correlation ($\circledcirc$) & Subtraction $-$\\
Bundling ($\oplus$) & Sum \& normalization & ---\\
Similarity ($\odot$) & Cosine similarity ($\text{cos}(\cdot,\cdot)$) & ---\\
\bottomrule
\end{tabular}
\end{table}

\subsubsection{Learning RPM Rules as an Assignment Problem.}
The core idea introduced in Learn-VRF is that the rules used in RPM can be framed in a fixed template which encompasses a series of binding and unbinding operations,
\begin{align}
    \label{eq:template}
    r = \left( \mathbf{c}_1 \circledast \mathbf{c}_2 \circledast \mathbf{c}_3 \right) \circledcirc \left( \mathbf{c}_4 \circledast \mathbf{c}_5 \circledast \mathbf{c}_6 \right),
\end{align}
where $\mathbf{c}_i$ represents a context panel $\mathbf{v}_a^{(i,j)}$ or the identity $\mathbf{e}$. 
In this setting, learning the rules of RPM can hence be interpreted as an assignment problem between VSA vectors and terms of Equation \ref{eq:template}.
To make it differentiable, Learn-VRF frames every term $\mathbf{c}_i$ as a convex combination over the VSA vectors of the context panels' attributes, augmented with the neutral element 
\begin{align}
    \label{eq:opti}
   \mathbf{c}_k = \sum_{\text{panels } (i,j)} w_k^{(i,j)} \cdot \mathbf{v}_a^{(i,j)} +  v_k\cdot \mathbf{e},
\end{align}
where the following constraints apply to the weights
\begin{align*}
   \sum_{\text{panels } (i,j)} w_k^{(i,j)} +  v_k = 1, \quad \quad 0 \leq w_k^{(i,j)} \leq 1 \, \forall i,j, \quad \quad 0 \leq v_{k} \leq 1 ,      \, \forall k.  
\end{align*}

\subsubsection{Executing and Selecting the Learned Rules.}
Inference with the learned rule set is a two-step process: an execution step (where all the rules are applied in parallel to the input) and a selection step (where a prediction for the missing panel is generated).
The application of each rule $r$ to an RPM example generates a tuple of three VSA vectors $(\hat{\mathbf{v}}_{a,r}^{(i,3)})^3_{i=1}$, which corresponds to the result of the rule execution on the three rows of the RPM matrix, together with a rule confidence value $s_r$.
The confidence value is computed as the sum of the cosine similarities between the predicted VSA vectors and  their respective ground-truth vector,
\begin{align}
\label{eq:conf}
    {s}_r =  \sum_{i=1}^3 \mathrm{cos}\left(\mathbf{v}_a^{(i,3)},  \hat{\mathbf{v}}_{a,r}^{(i,3)} \right).
\end{align}
During inference, the last term of the sum ($i=3$) is omitted, as the ground truth for the third row is unknown.

The answer is finally produced by taking a linear combination of the VSA vectors generated by executing all the rules, weighted by their respective confidence scores (normalized to a valid probability distribution using a softmax function).
More formally, if we define $\mathbf{s}=\left[ s_1, \dots, s_R \right]$ to be the concatenation of all rules' confidence score and $\hat{\mathbf{V}}_a^{(3,3)} = [ \hat{\mathbf{v}}_{a,1}^{(3,3)}, \dots,  \hat{\mathbf{v}}_{a,R}^{(3,3)} ]$ to be the concatenation of all rules' predictions for the missing panel, the final VSA vector predicted by the model for the attribute $a$ becomes
\begin{align}
\label{eq:softsel}
    \hat{\mathbf{v}}_a^{(3,3)} = \text{softmax}\left(\mathbf{s}\right) \cdot \hat{\mathbf{V}}_a^{(3,3)}. 
\end{align}
The use of the weighted combination can be understood as a \textit{soft selection} mechanism between rules and was found to be more effective compared to the \textit{hard selection} mechanism provided by sampling~\cite{hersche2023probabilistic}.

\section{Methods}
In this section, we present our \longname (\name) system.
An overview of \name is depicted in Figure~\ref{fig:carl}.
The framework aligns with the original Learn-VRF at a conceptual level, albeit with key adjustments that improve its expressiveness and boost its downstream performance on the I-RAVEN dataset.







\subsection{Learning Context-Augmented RPM Rules}
The soft-assignment problem presented in Equation \ref{eq:opti} is designed to assign each term $\mathbf{c}_i$ in Equation \ref{eq:template} with a fixed, absolute position in the $3\times 3$ RPM matrix.
For instance, the rule for arithmetic subtraction could be learned with one-hot assignment weights as $\hat{\mathbf{v}}_a^{(3,3)} = \mathbf{v}_a^{3,1} \circledcirc \mathbf{v}_a^{3,2}$.
A major limitation of this approach is that \emph{the rules cannot be shared across rows of the RPM matrix}. 
For example, the aforementioned arithmetic subtraction rule is valid only for the third row, but not for the first and second rows.

To overcome this limitation, Learn-VRF instantiates and learns three different rule sets (one per row) simultaneously.
During inference, the model leverages the first two sets to produce confidence values (Equation \ref{eq:conf}), which are then used to perform a soft selection of the output panels produced by the third rule set (Equation \ref{eq:softsel}).
While it was empirically shown to be effective, this implementation leaves the door open to different criticalities.
For instance, the model has no constraint on the functional equivalence between the three learned rule sets.
This renders the interpretability of the model sensibly harder and increases the likelihood of learning spurious correlations in the rule selection mechanism.
Furthermore, this formulation diminishes the model's versatility, as its design, tailored to the RPM context, cannot seamlessly transfer to other abstract reasoning tasks without a reconfiguration of its primary components.

\begin{figure}[t]
     \centering
     \begin{subfigure}[h]{0.3\textwidth}
         \centering
         \includegraphics[width=\textwidth]{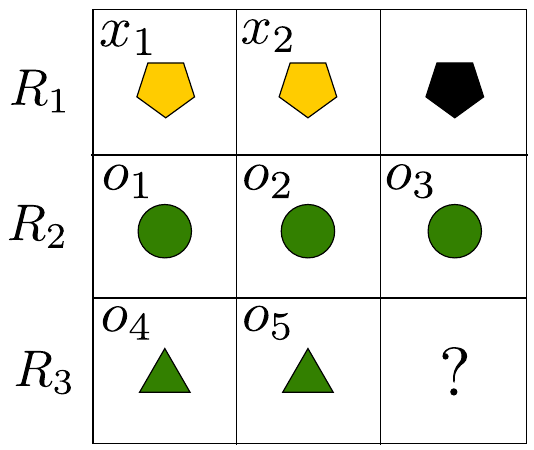}
         \caption{$X=\{R_1\}, O=\{R_2,R_3\}$}
         \label{fig:ov1}
     \end{subfigure}
     \hfill
     \begin{subfigure}[h]{0.3\textwidth}
         \centering
         \includegraphics[width=\textwidth]{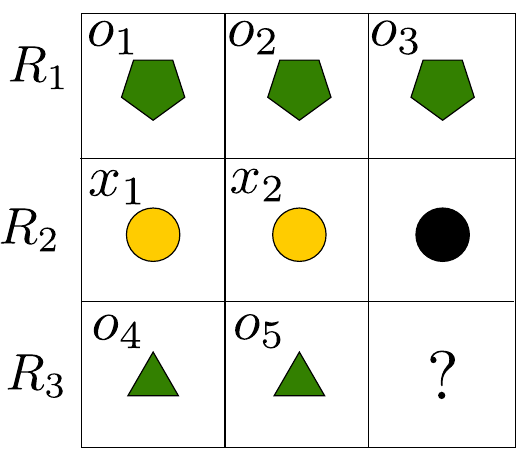}
         \caption{$X=\{R_2\}, O=\{R_1,R_3\}$}
         \label{fig:ov2}
     \end{subfigure}
     \hfill
     \begin{subfigure}[h]{0.3\textwidth}
         \centering
         \includegraphics[width=\textwidth]{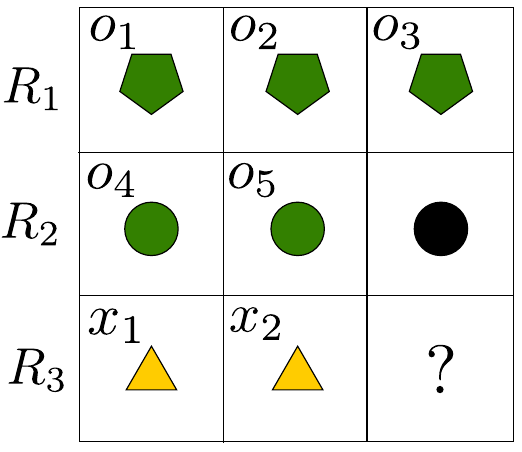}
         \caption{$X=\{R_3\}, O=\{R_1,R_2\}$}
         \label{fig:ov3}
     \end{subfigure}
        \caption{Visualization of current samples ($X=\{x_1,x_2\}$, in yellow) and context ($O=\{o_1,\dots,o_5\}$, in green) panels when predicting the third panel for different rows, namely the first row (left), second row (center) and third row (right). Black objects represent panels that are not used for the computation, while the question mark represents the unknown test panel, which is unavailable during inference.}
        \label{fig:overview}
\end{figure}

Motivated by these issues and by related works in cognitive sciences and psychology that argue for the importance of context in the solution of analogies for humans ~\cite{chalmers1992high,yozing1990context}
we propose a more general formulation of the soft-assignment problem which abstracts away the positional assignment and instead relies on the notion of \textit{context}.
We propose to rewrite Equation \ref{eq:opti} as
\begin{align}
    \label{eq:newopti}
   \mathbf{c}_k = \sum_{i=1}^I w_k^i \cdot \mathbf{x}_i  \cdot + \sum_{j=1}^J u_k^j \cdot \mathbf{o}_j  +  v_k\cdot \mathbf{e}.
\end{align}
Here, $\mathbf{X}=\{\mathbf{x}_1, \dots, \mathbf{x}_I\}$ is the set of attributes that define the current sample, that is, the description of the problem for which we infer a solution.
$\mathbf{O}=\{\mathbf{o}_1, \dots, \mathbf{o}_J\}$ is the set of attributes that define the context for that sample, that could be interpreted as a working memory from which additional information to infer the answer can be retrieved. 
In Equation \ref{eq:newopti}, $\mathbf{w},\mathbf{u},\mathbf{v}$ are the learned parameters and, as in Equation \ref{eq:opti}, they are subject to the following constraints:
\begin{small}
\begin{align*}
   \sum_{i=0}^I w_k^i + \sum_{j=0}^J w_k^j +  v_k = 1, \quad \quad 0 \leq w_k^i \leq 1 \, \forall i, \quad \quad 0 \leq u_k^j \leq 1 \, \forall j, \quad \quad 0 \leq v_{k} \leq 1, \, \forall k.
\end{align*}
\end{small}
Note that the notion of the current sample $X$ and its context $O$ depends on the row chosen for inference, as shown in Figure \ref{fig:overview}.

The new formulation does not lose expressiveness compared to Equation \ref{eq:opti}. 
While its terms are no longer tied to fixed positions in the RPM matrix, relative position information can still be preserved by keeping the order of the current and context samples consistent during training.
Both row-wise and column-wise relations can be correctly represented by the model.

Most importantly, the new context-aware formulation for the soft-selection allows to have a single set of rules shared across all the rows of the RPM.
Contrary to Learn-VRF, the model can now enforce functional equivalence between the rules used for selection and execution \textit{by construction}.
Additionally, the number of trainable parameters is reduced by $66\%$ compared to Learn-VRF.

In RPM, the number of current and context examples is equal to $I=2$ and $J=5$, respectively.
We do not consider $J=6$ context examples to ensure that the same rules can be shared across rows.
Otherwise, the model would fail when used to predict $R_1$ and $R_2$, since the panel in position $(3,3)$ is unknown.

\captionsetup{font=small} 
\captionsetup{font=normalsize}

\subsection{Improving Rule Selection through Template Generalization}
Compared to Learn-VRF, we increase the number of terms in the general rule template (Equation \ref{eq:template}) as
\begin{align}
    \label{eq:newtemp}
    r = \left( \mathbf{c}_1 \circledast \mathbf{c}_2 \circledast \mathbf{c}_3 \circledast \mathbf{c}_4 \circledast \mathbf{c}_5 \circledast \mathbf{c}_6 \right) \circledcirc \left( \mathbf{c}_7 \circledast \mathbf{c}_8 \circledast \mathbf{c}_9 \circledast \mathbf{c}_{10} \circledast \mathbf{c}_{11} \circledast \mathbf{c}_{12} \right).
\end{align}    
Increasing the representational power of the rule template opens up the possibility to learn more general---yet functionally equivalent--- formulations of the RPM rules, by also including ``validation'' terms in their structure, necessary in specific edge cases of the I-RAVEN dataset, such as Example \ref{ex:newterms}.
This bridges the gap between Learn-VRF, which could learn a set of rules that are perfect for execution but sub-optimal for selection, and other neuro-symbolic approaches~\cite{hersche2023neuro,PrAE_CVPR21}, which hard-coded an optimal rule set for selection and an optimal rule set for execution.
We define the \textit{optimality} of a rule set as follows.
\begin{definition}[Rule Set Optimality]
Consider a rule set $\mathcal{R}=\{r_i\}_{i=1}^R$ and an arbitrary RPM test $\mathbf{V}_a=(\mathbf{v}_a^{(1,1)},\dots,\mathbf{v}_a^{(3,2)})$. $\mathcal{R}$ is defined to be optimal for \textbf{execution} if $\, \exists \, r\in \mathcal{R}$ s.t. $\mathbf{v}_a^{(3,3)} = r\left( \mathbf{V}_a \right)$, and optimal for \textbf{selection} if the probability distribution  over $\mathcal{R}$ induced  by the selection mechanism (through confidence values $s_r$) concentrates all the probability on the correct rule.
\end{definition}
The importance of this distinction can be understood in the following example.

\begin{example}
\label{ex:newterms}
Consider the following RPM test example $V$, where different numbers correspond to different color attribute values,
\begin{align*}
{V} = 
\begin{matrix}
\begin{bmatrix}
\begin{tikzpicture}
    \matrix (magic) [matrix of nodes,ampersand replacement=\&] {
        9 \& 0 \& 9 \\
        6 \& 3 \& 9 \\
        3 \& 2 \& ? \\
    };
\end{tikzpicture}
\end{bmatrix}
\end{matrix}
\text{, with }
\begin{cases}
x_1=9, x_2=0, o_1=6, o_2=3, o_3=9 \text{ when } X=R_1\\
 x_1=6, x_2=3, o_1=9, o_2=0, o_3=9 \text{ when } X=R_2\\
 x_1=3, x_2=2, o_1=9, o_2=0, o_3=9 \text{ when } X=R_3
\end{cases}
\end{align*}
and a rule set $\mathcal{R}$ including the two rules that were proposed in Learn-VRF~\cite{hersche2023probabilistic} to solve \textit{arithmetic plus} (+) and \textit{distribute three} (d3), rewritten accordingly to our context-augmented formulation
\begin{gather*}
    \mathbf{v}^+ = \mathbf{x}_{1} \circledast \mathbf{x}_{2}  \equiv  {x}_{1} + {x}_{2} = v^+  \\
     \mathbf{v}^\text{d3} = \left( \mathbf{o}_{1} \circledast \mathbf{o}_{2} \circledast\mathbf{o}_{3} \right) \circledcirc \mathbf{x}_{1} \circledcirc \mathbf{x}_{2} \equiv   \left( {o}_{1} + {o}_{2} +{o}_{3} \right) - {x}_{1} - {x}_{2} = {v}^\text{d3}
\end{gather*}
where the equivalence between vector space and $\mathbb{R}$ is given by FPE.

Performing the selection using $\mathcal{R}$ can potentially result in a failure of the model for this RPM test.
In fact, we can see that both ${v}^+$ and ${v}^\text{d3}$ will produce the correct answer on the first two rows, and the confidence values (cosine similarity between the output and the true attribute) will be 1 for both.
As a result, the model will assign equal probabilities to both rules (even if only one of them is valid), rendering the probability of choosing the correct one equal to a coin toss.

Incorporating a validation term in the rule definition for \textit{distribute three} can, in this case,  solve the issue.
Consider the functionally equivalent rule
\begin{align*}{v}^\text{d3++}=\underbrace{\left( {o}_1 + {o}_2  + o_3 \right) - x_1 - x_2}_{\text{execution}} + \underbrace{\left( {o}_1 + {o}_2 + o_3 \right) - \left( {o}_1 + {o}_4 + {x}_1 \right)}_{\text{validation}}.
\end{align*}
The last two additional terms only cancel out when the sum of the elements of the first context row is equal to the sum of the elements of the first column, which is a property that is always verified for \textit{distribute three} but not for the other RPM rules.
Hence, using $ {v}^\text{d3++}$ instead of $ {v}^\text{d3}$ the model will be able to rule it out from the list of valid rules, correctly putting all the probability mass into the \textit{arithmetic plus} rule instead.
Note that learning $ {v}^\text{d3++}$ would not have been possible only with the 6 terms available in Equation \ref{eq:template}.
\end{example}

Therefore, we claim that Equation \ref{eq:newtemp}, even if still not optimal for selection, will increase the robustness of the model to RPM edge cases.
However, the increase in expressiveness also comes at a cost in terms of the number of trainable parameters, which scales linearly with the number of terms of Equations \ref{eq:newtemp}.

\subsection{Training Loss and other Implementation Aspects}
We follow the training recipe provided by Learn-VRF~\cite{hersche2023probabilistic}.
The training loss is defined as the inverse cosine similarity between the three predicted  panels  and their corresponding ground truth
\begin{align}
    \mathcal{L}= 1- \sum_{i=1}^3 \mathrm{cos}\left(\mathbf{v}_a^{(i,3)},  \hat{\mathbf{v}}_a^{(i,3)}\right).
\end{align}

As in Learn-VRF, we set the number of rules to $R=5$.
A single set of rules is instantiated and shared between all RPM attributes.
In previous works~\cite{hersche2023probabilistic}, the execution of the rules on the position and number attributes is performed in superposition, since either the number or the position attribute contributes to the generation of the answer.
However, the superposition requires a preliminary binding operation with (trainable) key vectors to avoid the binding problem~\cite{greff2020binding}.
As a result, vector arithmetic is no longer supported on these attributes.
We disentangle the two attributes, speculating that the trade-off between the additional noise introduced by the ``unused'' attribute, for which no rule is formally defined, and the increased computing accuracy will be significantly in favor of the latter.
Removing the superposition also allows us to remove the keys used by the binding, which were \textit{trainable parameters} in Learn-VRF and constituted the majority of parameters of the model (81\%).

\newpage

\section{Results}
\label{sec:results}

\subsection{In-distribution (ID) Results}
Table~\ref{tab:idresults} shows the \name's in-distribution downstream accuracy on I-RAVEN compared to a range of neuro-symbolic and connectionist baselines.
We present results for three different versions of \name: $\text{\name}_\text{progr}$, where the model's weights are manually programmed with RPM rules ($R=4$, since \textit{constant} can be considered as a special case of \textit{progression}), $\text{\name}_{\text{p}\mapsto\text{l}}$, where the model is initialized with the programmed rules and then trained with gradient descent, and $\text{\name}_\text{learn}$, where the rules are learned from scratch from data.

\begin{table}[b!]
\centering
\resizebox{\textwidth}{!}{
\begin{tabular}{ l c c l l l l l l l l} 
\toprule
\multicolumn{4}{c}{} & \multicolumn{7}{c}{I-RAVEN constellation accuracy (\%)} \\
\cline{5-11} 
Method & Approach&  Param. &  Avg. & C & 2x2 & 3x3 & L-R & U-D & O-IC & O-IG \\
\midrule

GPT-3 \cite{brown2020gpt3}& Selective  & 175\,b  & 86.5 & 86.4 & 83.2 & \textbf{81.8} & 83.4 & 84.6 & 92.8 & \textbf{93.0} \\
MLP \cite{hersche2023probabilistic}& Predictive & 300\,k& 87.1 & 97.6 & 87.1 & 61.8 & 99.4 & 99.4 & 98.7 & 65.6\\

SCL \cite{wu2020scl} & Selective &  961\,k & $84.3^{\pm 1.1 }$ &  $\mathbf{99.9^{\pm 0.0 }}$ & $68.9^{\pm1.9}$ & $43.0^{\pm 6.2}$ & $98.5^{\pm 2.9 }$ & $99.1^{\pm 1.5}$ & $97.7^{\pm 1.3 }$ & $82.6^{\pm 2.5 }$ \\
PrAE \cite{PrAE_CVPR21} & Predictive&n.a. & $71.1^{\pm 2.1 }$ &  $83.8^{\pm 3.4 }$ & $82.9^{\pm 3.3}$&$47.4^{\pm 3.2}$ & $94.8^{\pm 2.1 }$ & $94.8^{\pm 2.1 }$ & $56.6^{\pm 3.0 }$ & $37.4^{\pm 1.7 }$ 
\\
NVSA \cite{hersche2023neuro}& Predictive &n.a.& $88.1^{\pm 0.4 }$ &  $99.8^{\pm 0.2 }$ & $\mathbf{96.2^{\pm 0.8}}$ & $54.3^{\pm 3.2}$ & $\mathbf{100^{\pm 0.1 }}$ & $\mathbf{99.9^{\pm 0.1 }}$ & $\mathbf{99.6^{\pm 0.5 }}$ & $67.1^{\pm 0.4 }$\\

Learn-VRF \cite{hersche2023probabilistic}& Predictive  &  20\,k& $79.5^{\pm4.3}$ &$97.7^{\pm4.1}$ & $56.3^{\pm7.3}$ & $49.9^{\pm2.8}$ & $94.0^{\pm5.0}$ & $95.6^{\pm5.0}$ & $98.3^{\pm2.5}$ & $64.8^{\pm3.4}$ \\
\hline

$\text{\name}_\text{progr}$& Predictive &  n.a.& $92.4^{\pm 0.0}$ & $97.2^{\pm 0.0}$ & $84.9^{\pm 0.0}$ & $81.7^{\pm 0.0}$ & $97.8^{\pm 0.0}$ & $96.8^{\pm 0.0}$& $98.2^{\pm 0.0}$ & $90.0^{\pm 0.0}$\\
$\text{\name}_{\text{p}\mapsto\text{l}}$ & Predictive & 480&  $\mathbf{92.6^{\pm0.2}}$& $97.6^{\pm 0.0}$ & $84.7^{\pm 0.4}$  & $80.7^{\pm 0.4}$  & $98.5^{\pm 0.0}$  & $97.9^{\pm 0.0}$  & $98.5^{\pm 0.0}$  & $90.3^{\pm 0.3}$ \\
$\text{\name}_\text{learn}$& Predictive & 480&  $92.4^{\pm1.5}$& $98.4^{\pm 1.5}$ & $83.4^{\pm 1.6}$  & $80.0^{\pm 2.1}$  & $98.7^{\pm 1.2}$  & $98.4^{\pm 1.4}$  & $98.8^{\pm 1.2}$  & $89.4^{\pm 1.6}$ \\

\bottomrule
\end{tabular}
}
\vspace{0.3cm}
\caption{In-distribution accuracy (\%) on the I-RAVEN dataset. The results for PrAE and \name are obtained training only on the 2x2 constellation and testing on all the others. Among the baselines, we replicate Learn-VRF; the other results are taken from~\cite{hersche2023neuro}. The standard deviations are reported over 10 random seeds.} 
\label{tab:idresults}
\end{table}

\name achieves the best average accuracy on the in-distribution I-RAVEN dataset, improving the second-best result (NVSA~\cite{hersche2023neuro}, where the rules are hard-wired into the model) by almost $5\%$, while having orders of magnitude fewer parameters than any other baseline model.
\name also shows lower variance compared to the other baselines on almost every constellation.

Contrary to all the other methods (except for PrAE~\cite{PrAE_CVPR21}), \name is exclusively trained on the 2x2 constellation, effectively reducing the number of trained parameters and training samples by 85\% ($\sfrac{6}{7}$).
Its seamless adaptation to unseen constellations demonstrates the generality of the learned rules but prevents the model from outperforming the baselines in each single constellation.

\name produces close to perfect results on all the constellations without the position/number attribute (that is, C, L-R, U-C, and O-IC), strongly outperforming GPT-3 and PrAE.
On the other hand, its accuracy degrades for the constellations that include the position/number attribute (that is, 2x2, 3x3, and O-IG).
This degradation arises because of three specific rules on the position attribute: progression (corresponding to a circular bit-shifting operation), arithmetic plus (corresponding to the logical operation $a \lor b$), and arithmetic minus (corresponding to the logical operation $a \land \neg b$).
These rules, which operate at the granularity of \textit{objects}, cannot be easily captured by the model, which operates at the granularity of \textit{panels}.

Finally, comparing the three proposed versions of \name, it is interesting to observe that $\text{\name}_{\text{p}\mapsto\text{l}}$ outperforms both its fully-learned and fully-programmed equivalents.
The post-programming training allows to extend the knowledge of the model, rather than completely erasing it as shown in other settings~\cite{wu2023numerosity}, resulting in a monotonic increase in downstream accuracy.

\begin{table}[t!]
\caption{Ablation on the proposed improvements, as in-distribution accuracy (\%) on the I-RAVEN dataset. The ablations are row-wise incremental.} 
\label{tab:ablations}
\centering
\resizebox{\textwidth}{!}{
\begin{tabular}{ l  c c c c c c c c c } 
\toprule
Method   & C & 2x2 & 3x3 & L-R & U-D & O-IC & O-IG & Avg. & $\Delta$ \\
\midrule

Learn-VRF \cite{hersche2023probabilistic} & $97.7^{\pm4.1}$ & $56.3^{\pm7.3}$ & $49.9^{\pm2.8}$ & $94.0^{\pm5.0}$ & $95.6^{\pm5.0}$ & $98.3^{\pm2.5}$ & $64.8^{\pm3.4}$ & $79.5^{\pm4.3}$ & --\\

$\ $ p/n sup, 2x2 train & $92.1^{\pm6.5}$ & $78.5^{\pm5.4}$ & $76.1^{\pm5.3}$ & $92.4^{\pm6.8}$ & $92.3^{\pm6.6}$ & $95.1^{\pm4.2}$ & $86.6^{\pm3.4}$ & $87.6^{\pm5.5}$ & $\mathbf{+8.1}$\\
$\ $ context-awareness & $95.2^{\pm4.2}$ & $81.9^{\pm3.4}$ & $78.9^{\pm3.6}$ &  $95.1^{\pm4.4}$ & $94.9^{\pm4.5}$ & $96.6^{\pm2.7}$ & $88.8^{\pm2.1}$ & $90.2^{\pm3.6}$ & $+2.6$\\
$\ $ 12 terms (\name) & $\mathbf{98.4^{\pm 1.5}}$ & $\mathbf{83.4^{\pm 1.6}}$  & $\mathbf{80.0^{\pm 2.1}}$  & $\mathbf{98.7^{\pm 1.2}}$  & $\mathbf
{98.4^{\pm 1.4}}$  & $\mathbf{98.8^{\pm 1.2}}$  & $\mathbf{89.4^{\pm 1.6}}$ & $\mathbf{92.4^{\pm1.5}}$& $+2.2$\\

\bottomrule
\end{tabular}
}
\end{table}

Table~\ref{tab:ablations} reports a thorough ablation on the novelties introduced in our framework.
We can observe that the biggest contribution comes from removing the position/number superposition ($+8.1\%$), which consistently increases the performance on the constellations involving these attributes.
However, in all the other constellations we observe a consistent drop in accuracy, due to the evaluation of the model in the ``transfer'' setting (trained on 2x2, evaluated on all the others).
This drop is compensated by the two novel components of the model, the context-awareness and the generalized rule template, which allow \name to match and outperform Learn-VRF on every constellation.
Interestingly, they also contribute to reduce the variance of the results, suggesting that the two improvements might be increasing the model invariance to weight initialization.

\subsection{Out-of-distribution (OOD) Results}
We validate \name's out-of-distribution generalization capabilities by following the same recipe proposed by Learn-VRF~\cite{hersche2023probabilistic}: the model is trained on a subset of the rule-attributes pairs of the center constellation, and evaluated on its complement.
As shown in Table~\ref{tab:oodresults}, \name matches the OOD performance previously showed by Learn-VRF, attaining perfect accuracy on almost every unseen rule-attribute pair in the center constellation. 

\begin{table}[t!]
\caption{OOD accuracy (\%) on unseen rule-attribute pairs on I-RAVEN.} 
\label{tab:oodresults}
\resizebox{\textwidth}{!}{
\begin{threeparttable}
\begin{tabular}{l l l l l l l l l l l l}
\toprule
        & \multicolumn{3}{c}{Type} & \multicolumn{4}{c}{Size}  & \multicolumn{4}{c}{Color} \\
        \cmidrule(r){2-4} \cmidrule(r){5-8} \cmidrule(r){9-12}
        & Const.      & Progr.      & Dist.3      & Const.      & Progr.      & Dist.3 & Arith.    & Const.     & Progr.      & Dist.3 & Arith.  \\
\hline
GPT-3 \cite{hersche2023probabilistic} & 88.5 & 86.0 & 88.6  & 93.6 &93.2 & 92.6 & 71.6 & 94.2 & 94.7 & 94.3 & 65.8\\
MLP baseline \cite{hersche2023probabilistic} & 14.8 & 14.9 & 30.2 & 22.8 & 75.0 & 47.0 & 46.6 & 56.3 & 60.5 & 44.4 & 48.9\\
Learn-VRF \cite{hersche2023probabilistic} & 100 & 100 & 99.7 & 100 & 100 & 99.8 & 99.8 & 100 & 98.8 & 100 & 100 \\
\hline
\name &100 & 98.6 & 99.7 & 100 & 100 & 99.6 & 99.6 & 100 & 100 & 100 & 99.8 \\
\bottomrule
\end{tabular}
\end{threeparttable}
}
\end{table}

\newpage
\section{Conclusions and Future Work} 
In this work, we proposed the \longname (\name), a model built on top of Learn-VRF~\cite{hersche2023probabilistic} to enhance its downstream accuracy, interpretability, model size, and coherence.
We conducted evaluations on the I-RAVEN dataset, demonstrating significant performance improvements compared to a diverse array of neuro-symbolic and connectionist baselines, including large language models, across both ID and OOD data.
Furthermore, we presented empirical results on the programmability of the model and the generalization across different constellations of the I-RAVEN dataset.

A potential avenue for future research involves addressing the remaining challenge posed by this dataset, specifically, the development of suitable representations to allow the learnability of arithmetic and progression rules on the position attributes. These rules are currently impossible to learn for the model and would allow it to attain perfect accuracy on I-RAVEN.
Additionally, extending the evaluation of \name to include other reasoning benchmarks, such as ARC~\cite{chollet2019measure}, also represents a promising direction for further investigation.
While the scope of this work was mostly focused on developing and studying a prototype that could solve RPM, the proposed general formulation of the problem could transfer to other settings that require learning relations and analogies from data. Furthermore, as our experiments on learning of top programmed knowledge show, our framework holds potential for application in scenarios where only a partial knowledge of the dynamics is available, and it is necessary to discover and build new knowledge on top of it.

\subsubsection{Acknowledgments} This work is supported by the Swiss National Science Foundation (SNF), grant 200800.
\newpage

%
%
\bibliographystyle{splncs04}

\end{document}